\pdfoutput=1

\documentclass[11pt]{article}

\usepackage[final]{acl}

\usepackage{times}
\usepackage{latexsym}

\usepackage[T1]{fontenc}

\usepackage[utf8]{inputenc}

\usepackage{microtype}

\usepackage{inconsolata}

\usepackage{graphicx}
\usepackage{microtype}
\usepackage{graphicx}
\usepackage{subfig}
\usepackage{booktabs} 

\usepackage{algorithm}
\usepackage{algpseudocode}
\usepackage{enumitem}
\usepackage{tabularx}
\usepackage{adjustbox}
\usepackage{longtable}
\usepackage{caption}
\usepackage[utf8]{inputenc}

\usepackage{amsmath, amssymb}
\usepackage{tcolorbox}
\usepackage{graphicx} 
\usepackage{float}
\usepackage{rotating}
\usepackage{microtype}
\usepackage{hyperref}
\usepackage{url}
\usepackage{booktabs}
\definecolor{darkblue}{rgb}{0, 0, 0.5}
\hypersetup{colorlinks=true, citecolor=darkblue, linkcolor=darkblue, urlcolor=darkblue}

\title{Can Multi-turn Self-refined Single Agent LMs with Retrieval Solve Hard Coding Problems?}

\author{
  \textbf{Md Tanzib Hosain\textsuperscript{1,3,}\thanks{Work done while working as a remote RA at QCRI.}},
  \textbf{Md Kishor Morol\textsuperscript{2,3}}\\
  \textsuperscript{1}American International University-Bangladesh,
  \textsuperscript{2}Cornell University\\ \textsuperscript{3}ELITE Research Lab\\
 \texttt{20-42737-1@student.aiub.edu, mmorol@cornell.edu}
}

\begin{document}
\maketitle
\begin{abstract}
Among the hardest tasks for humans are those found in competitive programming where problems require sophisticated algorithmic thinking, puzzle solving, and the creation of effective code. As a domain to assess language models (LMs), it has not received enough attention, though. This study presents the ICPC benchmark, which consists of 254 international collegiate programming contest (ICPC) tasks. Each problem includes official analysis, reference code, and sample, high-quality unit, and hidden tests. We are able to develop and evaluate a variety of LM inference techniques for competitive programming with these resources. With zero-shot chain-of-thought prompting, we find that o1 only achieves a 19.1\% pass@1 solve rate. With our best inference technique, which combines multi-turn self-judge with reflection and retrieval over episodic information, raises this to 42.2\%. Furthermore, we conduct a new human-in-the-loop investigation to gain a deeper understanding of the remaining difficulties. Surprisingly, we discover that o1 can solve 17 out of 18 problems that were previously unsolvable by any model or technique with just a few specific instructions. A footstep toward LMs with grounded, imaginative, and algorithmic thinking is provided by our quantitative findings and qualitative research. We open-source our code and data at https://github.com/kraritt/zolve.
\end{abstract}

\section{Introduction}
A crucial area for assessing and implementing language models (LMs) is code generation. However, several well-known coding benchmarks, including HumanEval \cite{chen2021evaluating} and MBPP \cite{austin2021program}, have become saturated with solve rates above 90\% due to the scaling of LMs and the development of new inference techniques \cite{chen2023teaching, shinn2024reflexion, wei2022chain, zhou2022docprompting}. We require more difficult benchmarks that highlight the shortcomings of current models, inference techniques and offer practical instincts for enhancing LM's algorithmic reasoning in order to spur additional advancement. Since competitive programming where problems are intended to rigorously assess human reasoning skills in difficult circumstances and the development of innovative algorithms, it is a perfect fit for this endeavor. To thoroughly assess algorithmic reasoning, prior investigations of competitive programming, however, have either lacked full unit test suites, problem analysis, or sufficient problem variety \cite{jain2024livecodebench, li2022competition, hendrycks2021measuring}.

With 254 difficult competitive programming tasks from previous ICPC (including regional, continental, world final, etc.) contests, we provide a meticulously designed coding benchmark. As well as some sample tuples of inputs, outputs, and explanations, each challenge outlines a job to be completed in a made-up situation. Solving these problems require for both innovative and grounded thinking in addition to a broad variety of mathematical, computational, and common-sense expertise. With using zero-shot chain-of-thought prompting, even the best o1 only achieves a 19.1\% pass@1 solution rate. Apart from that, in order to investigate more sophisticated inference-time techniques for competitive programming, our benchmark also gathers official analysis, reference code solutions, and excellent unit and hidden tests for every problem, as well as the relevant teaching materials in the form of competition programming textbooks. Using these resources, we develop a variety of baseline techniques based on take-a-deep-breath prompt \cite{yang2024zhongjing}, brainstorm then select \cite{summers2023brainstorm}, zero-shot-CoT \cite{kojima2022large}, LLM Stimuli \cite{li2023large}, self-reflection \cite{shinn2024reflexion}, few-shot prompting \cite{brown2020language} and retrieval augmented generation- semantic and episodic retrieval \cite{su2024bright, gao2023retrieval, shypula2023learning}, and their combinations.

We discover that multi-turn self-judge single agent LMs with retrieval over comparable problems and solutions together with self-reflection increases performance by 120.94\% with respective to o1's zero-shot solve rate. Moreover, we conduct a unique human investigation to better understand the limitations and promise of LM reasoning toward competitive programming. In this study, humans engage with LMs in a conversational "tutoring" setup by pointing out errors and providing only a few tips. Interestingly, when we use a human-in-the-loop configuration, o1 solves 17 out of 18 tasks that can ever answer using any inference techniques. This suggests that stronger LMs may eventually be able to include high-quality input, that new techniques for producing such human-level corrective feedback must be developed, and the appropriate criterion for assessing model capabilities beyond the too stringent execution success should be reconsidered.

We require just black-box access to language model generations; no model-internal information, like as likelihoods or gradients, is required. We employ the same technique and prompt templates for all of our tasks.  This makes it possible to apply our approach with popular public models that provide interfaces. Additionally, further model generation enhancements like prompt engineering, self-reflection, or retrieval, are orthogonal to the approach.

In summary, the contributions of our work are provided in the following. At first, the benchmark based on contest programming that includes excellent unit and hidden test cases, problem analysis, and supplementary materials is the ICPC benchmark, which we propose. After that, we develop and evaluate several LM inference techniques for contest programming. Later, we provide a unique method that uses a multi-turn self-judge single-agent LMs with retrieval process to increase the reasoning of modern language models. Our findings show that multi-turn self-judge single-agent LMs with retrieval and self-reflection together can significantly improve performance. Finally, we combine automated tests based on execution success with a new human-in-the-loop research to describe the strengths and weaknesses of LMs for contest programming. Latent differences across models are revealed when we discover that only some models are able to correctly integrate feedback.

\section{Related Work}
\subsection{Problem Solving Coding Benchmarks} Numerous studies have examined language model performance on basic program synthesis \cite{zan2022large, austin2021program, chen2021evaluating, yu2018spider} and HumanEval—the industry standard for evaluating new models on code synthesis. But with the help of inference techniques, existing models can tackle HumanEval problems with a 94\% success rate \cite{zhou2023language}. This suggests that more challenging, intricate and self-contained coding challenges are required to test the limits of code reasoning. Thus, competitive programming questions have been suggested as a more challenging assessment metric. The majority of these tasks originate from online resources like Topcoder, LeetCode, Codeforces, Atcoder and others \cite{jain2024livecodebench, huang2023competition, li2023taco, li2022competition, hendrycks2021measuring}. Still, a considerable number of these challenges are only described symbolically and lack thorough test cases that define correctness and quality problem evaluations. The model's capacity to use creative reasoning in grounded task environments—a critical skill of well-rounded reasoners—is thus only marginally assessed.


\subsection{Inference Time Techniques}
According to \cite{chen2023teaching, gao2023retrieval, madaan2024self, shinn2024reflexion, zhou2023language, le2022coderl, yao2022react, zhou2023language}, inference time methods have demonstrated notable success in enhancing reasoning abilities by conditioning generations on environment feedback, task-specific knowledge, natural language reflections, and planned summaries. Nevertheless, only basic program synthesis tasks like HumanEval and MBPP have utilized their usefulness on code domains thus far \cite{austin2021program, chen2021evaluating}. In this study, we also discuss how well they perform in a far more challenging domain: competitive programming. We also draw inspiration for our retrieval augmented generation implementation from classical case-based reasoning literature \cite{aamodt1994case, schank1983dynamic} and cognitive architectures for human reasoning \cite{sumers2023cognitive}, which reflect the kinds of information that people find helpful in solving problems.

\subsection{Human Agent Interaction (HAI)}
Agent learning via human-provided feedback under synthetic tasks is examined by \cite{sumers2022talk}. The purpose of \cite{macina2023mathdial} is to offer a set of tutoring guidelines for successfully including LMs in conversation problem solving. In order to assess the models' capacity to react to feedback, we use a set of interaction rulesets from \cite{shi2024can}.

\section{Setup}
\subsection{Benchmarks}

\begin{table}[htbp]
\scriptsize
\centering
\caption{Problem count based on contest venue. 'WF' and 'CF' denote World and Continental Finals, respectively.}
\begin{tabular}{p{5cm}p{2.2cm}}
\hline
\textbf{Category} & \textbf{Problems\#} \\ \hline
WF \& CF & 167\\ \hline
Regional & 87\\ \hline
\textbf{Total} & \textbf{254}\\ \hline
\end{tabular}
\label{tab8}
\end{table}

From previous ICPC coding competitions, because of lacking strong co-relation with reasoning problem standards (extreme simple problems) we filtered out some problems and finally 254 expert-written, superior competitive programming tasks make up the ICPC benchmark, presented in Figure \ref{tab8} (For detail selection see Appendix \ref{tab1030}). An official human-written problem analysis stating the solution in detail with corresponding C++ code, some unit tests (sample and some synthesized tests) and hidden tests (synthesized tests) confirming solution correctness, time and memory limits confirming solution complexity and a problem description with instructions for reading and writing from standard input and output comprise each problem. Synthesized tests were produced from problem constraints with potential edge cases discussed in the official editorials and validated against official solutions to ensure correctness. This approach is standard in competitive programming research, mitigating reliance on public test cases \cite{schafer2023empirical}. A model is provided with the problem description, time and memory constraints, and any samples and synthesized tests as unit tests that are available. After that, the model retrieves related reference documents and using that as episodic knowledge (see in Section \ref{basel}) the model must provide a code solution that the same model judge (self-judge) judges and accepts if it enforces correctness and the intended asymptotic efficiency by yielding the predicted results on all unit tests (in this part, we selected the synthesized tests which don't exist in the hidden test cases) within the specified bounds and the process will terminate. In case the code fails on the unit tests, the whole process will repeat again until convergence or reach into the specified iteration (we found that $i = 2$ is ideal for o1 in this scenario--shown in Table \ref{tabiter}). After that the solution will execute against the hidden tests to get the final pass/fail results. A custom HTML5 parser is used to gather 254 tasks\footnote{\href{https://icpc.global/}{https://icpc.global/}} that explain contest materials. Regular expressions are then used to extract time and memory limits from problem descriptions. We choose 254 competitive programming tasks with complete problem analyses to aid in the creation of rich inference-time techniques and assessments. We parse a ground truth standalone C++ code snippet and an English-only analysis devoid of code for episodic knowledge retrieval. We ask GPT-4 to convert the code to C++ for tasks when C++ code is not accessible and we confirm that all code solutions pass hidden tests on the specified restrictions.

\subsection{Baselines}\label{basel}
\begin{figure}[htbp]
\centering
\includegraphics[width=.5\textwidth]{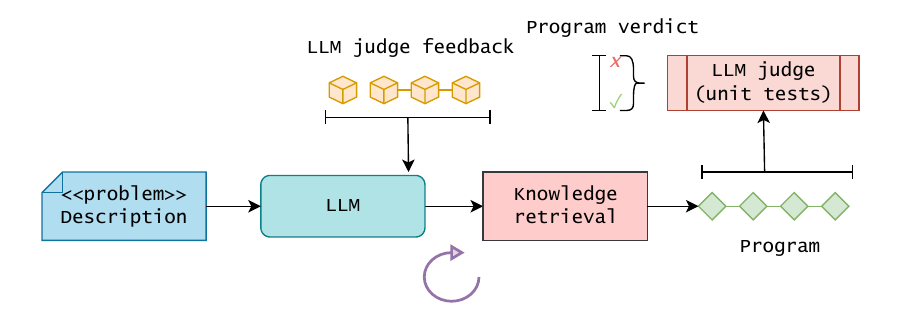} 
\caption{Framework architecture with Knowledge retrieval and Self-reflection.}
\label{fig2}
\end{figure}

We test a number of prompting and inference time strategies, including the take-a-deep-breath prompt \cite{yang2024zhongjing}, brainstorm then select \cite{summers2023brainstorm}, zero-shot-CoT \cite{kojima2022large}, LLM Stimuli \cite{li2023large}, self-reflection \cite{shinn2024reflexion}, few-shot prompting \cite{brown2020language} and retrieval augmented generation- semantic\footnote{As our resource, we utilize the Algorithms for Competitive Programming textbook, which includes chapters on algorithmic principles written by humans. \newline \href{https://cp-algorithms.com/}{https://cp-algorithms.com/}} and episodic retrieval \cite{su2024bright, gao2023retrieval, shypula2023learning}. As no single prompt performs better than the others (Table \ref{tab9}), we choose the episodic retrieval with reflection prompt in our single-agent LMs framework (Figure \ref{fig2}). Furthermore, to fully explore the potential of retrieval on the comparatively small dataset, we simulate a setup in which the model has seen every other problem in the ICPC set aside from the one it is currently solving. This is done by simulating an n-fold evaluation that presents one problem at a time. Although we get comparable results with a more traditional train-test split, as detailed in Section \ref{perf-met}. Concatenating the problem description, solution and C++ solution code for each seen problem creates documents that may be retrieved. After adjusting for the number of problems to retrieve, $p$, we determine that $p$ = 2 is ideal for o1. As pass@1 performance was declining, we decided not to try resampling for larger amounts of $p$ in order to save budget. As a result, we publish these values (Table \ref{tab6}).

\subsection{Metric}
We use every method that has a Pass@1 evaluation and the methods from \cite{shi2024can} for self-reflection and episodic retrieval, and we only give the models the execution outcomes of the exposed unit test cases. Fundamental studies were done using GPT-4, GPT-4o and o1 with some open source models tested in zero-shot setting only.

\section{Results}
\subsection{Performance Baselines}
\begin{table}[htbp]
\centering
\scriptsize
\caption{Pass@1 performances of various models for zero-shot problem-solving configuration.}
\begin{tabular}{p{3.6cm}p{3.6cm}}
\hline
\textbf{Model} & \textbf{Pass@1} \\ \hline
gpt-4 & 7.3 \\ \hline
claude-3.5-sonnet & 14.1 \\ \hline
gpt-4o & 14.2\\ \hline
qwen2.5-coder & 14.8\\\hline
athene-v2-chat & 16.4 \\ \hline
deepSeek-v3-chat & 17.6\\ \hline
gemini-exp & 18.3\\ \hline
\textbf{o1} & \textbf{19.1}\\ \hline
\end{tabular}
\label{tab11}
\end{table}
As a starting point, we assess the zero-shot performance of models that represent the state-of-the-art coding performance, such as GPT-4 (gpt-4-0613), GPT-4o (gpt-4o-2024-11-20), o1 (o1-2024-12-17), Claude-3.5-Sonnet (claude-3.5-sonnet-20240620), Gemini-Exp (gemini-exp-1206), Athene-V2-Chat (athene-v2-chat-72b), DeepSeek-V3-Chat, and Qwen2.5-Coder (qwen2.5-coder-32b-instruct) \cite{achiam2023gpt, team2024gemini, liu2024deepseek, hui2024qwen2}. Table \ref{tab11} provides an overview of this. If not otherwise noted, models were given chain-of-thought prompts \cite{wei2022chain}; the complete prompts are shown in Appendix \ref{prompt}. In accordance with earlier studies on competitive programming \cite{li2022competition, hendrycks2021measuring}, we mainly use the unbiased pass@n metric as specified in \cite{chen2021evaluating}. For that, we discover that compilation errors are not the primary cause of any model defects (see Section \ref{error-break}). This at least demonstrates that models are successful in producing syntactically sound code and points to more complex problems in generations, including miscommunications.

\subsection{Performance Benchmarks}\label{perf-met}
\begin{table*}[htbp]
\scriptsize
\centering
\caption{Pass@1 performances for various problem-solving configurations.}
\begin{tabular}{p{12.5cm}ccc}
\hline
\textbf{Inference technique} & \multicolumn{3}{c}{\textbf{Model}} \\ \cline{2-4}
 & gpt-4 & gpt-4o & o1 \\ \hline
zero\_shot                                & 7.3  & 14.2 & 19.1 \\ \hline
brainstorm\_then\_select                  & 8.6  & 16.9 & 21.7 \\ \hline
few\_shot                                 & 10.1 & 19.4 & 24.2 \\ \hline
self\_reflection                          & 11.3 & 20.6 & 25.4 \\ \hline
semantic\_retrieval                       & 12.4 & 22.1 & 27.3 \\ \hline
semantic\_retrieval + self\_reflection    & 12.8 & 22.5 & 28.1 \\ \hline
episodic\_retrieval                       & 13.2 & 23.3 & 29.0 \\ \hline
semantic\_retrieval + episodic\_retrieval  & 14.5 & 24.4 & 29.8 \\ \hline
semantic\_retrieval + episodic\_retrieval + self\_reflection & 16.4 & 27.1 & 33.2 \\ \hline
\textbf{episodic\_retrieval + self\_reflection} & \textbf{24.3} & \textbf{38.4} & \textbf{42.2} \\ \hline
\end{tabular}
\label{tab9}
\end{table*}

\begin{table}[htbp]
\scriptsize
\centering
\caption{Pass@1 performances when compared to our leave-one-out episodic retrieval situation, the outcomes of a normal train-test split are comparable across inference-time approaches.}
\begin{tabular}{p{4.75cm}ccc}
\hline
\textbf{Inference technique} & \multicolumn{3}{c}{\textbf{Model}} \\ \cline{2-4}
 & gpt-4 & gpt-4o & o1 \\ \hline
episodic\_retrieval                       & 10.9 & 18.6 & 22.7 \\ \hline
self\_reflection                          & 11.1 & 20.4 & 24.2 \\ \hline
\textbf{episodic\_retrieval + self\_reflection} & \textbf{21.3} & \textbf{33.8} & \textbf{35.4} \\ \hline
\end{tabular}
\label{tab7}
\end{table}

\begin{table}[htbp]
\scriptsize
\centering
\caption{o1 hyperparameter tuning on the number of problems to retrieve for episodic retrieval.}
\begin{tabular}{p{3.6cm}p{3.6cm}}
\hline
\textbf{Problems} & \textbf{Pass@1} \\ \hline
$p$ = 1 & 28.1 \\ \hline
\textbf{$p$ = 2} & \textbf{29.0} \\ \hline
$p$ = 3 & 28.4 \\ \hline
\end{tabular}
\label{tab6}
\end{table}

\begin{table}[htbp]
\scriptsize
\centering
\caption{o1 iteration tuning on the number of iterations for self-reflection. Without any reflection, the solve rate is $i$ = 0. We see that after 2 repetitions, solve rates nearly stay the same.}
\begin{tabular}{p{3.6cm}p{3.6cm}}
\hline
\textbf{Iterations} & \textbf{Pass@1} \\ \hline
$i$ = 0 & 21.3 \\ \hline
$i$ = 1 & 23.8 \\ \hline
\textbf{$i$ = 2} & \textbf{25.6} \\ \hline
$i$ = 3 & 25.4 \\ \hline
\end{tabular}
\label{tabiter}
\end{table}

Aligning with \cite{shi2024can, shinn2024reflexion, chen2023teaching}, we discover that stronger models have the emergent quality of being able to self-reflect successfully. Nevertheless, both episodic and semantic retrieval remain efficient; in fact, episodic retrieval even makes GPT-4o come close to o1's zero-shot performance (Table \ref{tab9}). This is probably due to the fact that self-reflection depends on the internal model's capacity to interpret binary, sparse reward signals. Conversely, retrieval enables models to make use of pre-existing logic and code fragments, necessitating less inherent model capabilities.
Thus, our results support \cite{li2023explaining}, which found that LMs are able to comprehend competitive programming solutions that are far more sophisticated than they are able to generate. Furthermore, combining episodic retrieval with reflection allows it to reach new heights, but not with semantic retrieval. The additional knowledge offered by our implementation of semantic retrieval trades off against its extended contexts, which existing LLMs are known to struggle with \cite{liu2024lost, shi2024can}. This offers one explanation for why combining the two might result in decreased performance.

Furthermore, instead of the model crucially interacting with the retrieved information itself, the opposing theory for retrieval success holds that adding obtained answers enhances memorizing effects for the problem under evaluation. To check for this, we eliminate crucial portions of the recovered solutions and see notable performance decreases. The created and officially published answers also do not significantly overlap, according to qualitative examination. Section \ref{ablation} contains the experiment specifics.

Additionally, for maximizing the impact of retrieval on the comparatively short dataset at hand, our episodic retrieval assessment setup entails presenting one problem at a time that is retrieves from the solutions of all other test problems, as explained in Section \ref{basel}. Given how independent problems are and how little solution logic even problems with the same method type share, we anticipate that this will not result in any notable dataset leaking across evaluations. We did, however, rerun most of the inference-time methods against a more conventional train-test split arrangement. The conventional split, train size = 200, test size = 54 produces comparable results with somewhat lower retrieval efficacy, as seen in Table \ref{tab7}. This is due to the fact that fewer problems are retrieved overall, which results in a generally lower level of problem similarity between the problems that are recovered and the ones that are being addressed at the moment. Moreover, we recover the same optimal values as the leave-one-out configuration by re-tuning the number of recovered passages solely on this train set.

\subsection{Performance HAI}
\begin{figure}[htbp]
\centering
\includegraphics[width=.5\textwidth]{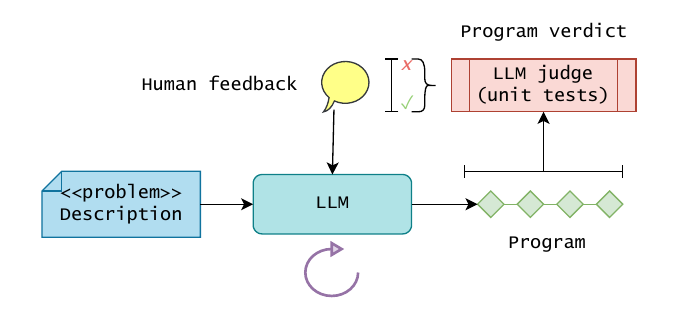} 
\caption{Framework architecture with integrating HAI.}
\label{fig4}
\end{figure}

\begin{table}[htbp]
\scriptsize
\centering
\caption{Feedback is integrated into o1's HAI interactive setting. (Final solve rate would be highly dependent on the problem-solving strength of the human performing the interactions with the models. For this case study, the participants who participated in this interaction module have Codeforces rating > 2500.)}
\begin{tabular}{p{3.6cm}p{3.6cm}}
\hline
\textbf{Model} & \textbf{Final solve rate} \\ \hline
gpt-4 & 0 \\ \hline
gpt-4o & 0 \\ \hline
o1 & 0 \\ \hline
\textbf{o1 + interact} & \textbf{94.4}\\ \hline
\end{tabular}
\label{tab10}
\end{table}
We discovered a broad range of model error distributions in benchmark assessments, ranging from minor off-by-one implementation problems to severe misconceptions. We conduct a human research using an interactive tutoring to further investigate how close a model is to resolving a particular task (Figure \ref{fig4}). Remarkably, we discover that the human-in-the-loop approach improves o1 performance from 0\% to 94.4\% (Table \ref{tab10}), 17 problems solved on a small set of 18 problems on which GPT-4, GPT-4o and o1 reach zero pass rate using all of the aforementioned inference-time methods, but does not improve GPT-4 and GPT-4o performance from 0\%. When two models fail on a particular problem, one may be one adjustment away from a completely perfect solution, while the other may have a basic misunderstanding of the problem scenario. These human-in-the-loop results demonstrate that the solve rate may not fully represent the capabilities of models. This encourages improved measures for assessment that go beyond execution success, pass@n. As an alternative interpretation of our findings, it is possible that human-level corrective feedback might open more thinking abilities in o1, underscoring the need for improved techniques to produce such feedback. Appendix \ref{hai-path} contains a scenario of the interaction pathway.

\subsection{Ablation Test}\label{ablation}
\begin{table}[htbp]
\scriptsize
\centering
\caption{Performance on various retrieval query ablations.}
\begin{tabular}{p{6cm}p{1.2cm}}
\hline
\textbf{Query} & \textbf{Pass@1} \\ \hline
problem\_description & 28.5 \\ \hline
\textbf{problem\_description + proposed\_code\_solution} & \textbf{29.0} \\ \hline
problem\_description + proposed\_solution + code\_solution & 29.8 \\ \hline
\end{tabular}
\label{tab4}
\end{table}

\begin{table}[htbp]
\scriptsize
\centering
\caption{Performance on various episodic retrieval ablations.}
\begin{tabular}{p{5cm}p{2.2cm}}
\hline
\textbf{Retrieval} & \textbf{of max performance} \\ \hline
problem\_description + code + solution & 100.0 \\ \hline
\textbf{problem\_description} & \textbf{2.3} \\ \hline
\end{tabular}
\label{tab5}
\end{table}

For the ICPC problemset, we do ablation test on various prompts in order to establish the parameters for the primary experiments.

Apart from that, in the investigation on how the prompts impact problem-solving in a conversation, we create a variety of specific prompts for our suggested self-feedback single agent with retrieval framework. Appendix \ref{prompt} incorporates the prompt designs and report the findings, identifying the prompt as the primary prompt for more research.

According to ablations on retrieval queries, the best retrieval queries make use of both the current problem description and a first solution attempt that includes code and an explanation. This makes it possible to accurately obtain pertinent algorithm descriptions from the underlying retrieval corpus, as retrieval over algorithmic keywords is not possible when only the issue descriptions are used. Since our local judge has not seen this first generation, we do not consider it an effort. For that, we found in Table \ref{tab4}, the majority of retrieval queries, in general, are rather effective; nevertheless, the best results are obtained by combining code proposes and proposed solutions, as this enables the greatest possible matching of pertinent keywords across the compared documents. Applying ablations to the corpora in Table \ref{tab5}, we tackle memorizing. If retrieving problem solutions was causing people to recite previously learned answers to the present problem, then eliminating important components of the obtained solutions would not lessen this impact. But we discover that it does: using only the problem description preserves just 2.3\% of the performance, indicating that models are actually using the context-provided reasoning of related problems to guide their generations.

\section{Errors}\label{error-break}
\begin{table}[htbp]
\scriptsize
\centering
\caption{Error distributions of episodic\_retrieval + self\_reflection, in \%. TLE indicates time limit exceeded, and MLE
Indicates memory limit exceeded. 'Other' generally represents errors stemming from
models outputting incorrectly formatted code.}
\label{taberror}
\begin{adjustbox}{max width=\linewidth}
\begin{tabular}{lccccc}
\hline
\textbf{Model} & \textbf{Wrong Ans.} & \textbf{TLE} & \textbf{MLE} & \textbf{Runtime} & \textbf{Syntax + Other} \\
\hline
gpt-4 & 58.81 & 5.33 & 0 & 10.16 & 1.38 \\
gpt-4o & 28.95 & 25.06 & 0 & 6.83 & 0.77 \\
o1 & 27.87 & 23.56 & 0 & 5.78 & 0.59 \\
\hline
\end{tabular}
\end{adjustbox}
\end{table}

Table \ref{taberror} indicates on where models are trading raw speed for more profound reasoning capabilities. While gpt-4 provides rapid but often incorrect solutions, gpt-4o and o1 engage in a more computationally expensive process that yields correct answers far more frequently. The o1 model establishes itself as the superior agent in this analysis, demonstrating marginal but consistent gains over gpt-4o in both correctness and efficiency. Future work should investigate methods to mitigate the high computational cost (TLEs) of these advanced models without compromising their newfound accuracy, perhaps through optimized algorithms or more efficient self-reflection triggers.

\section{Results Analysis}
\begin{figure}[h!]
    \centering
    \begin{minipage}[t]{0.32\textwidth}
        \centering
        \includegraphics[width=\textwidth]{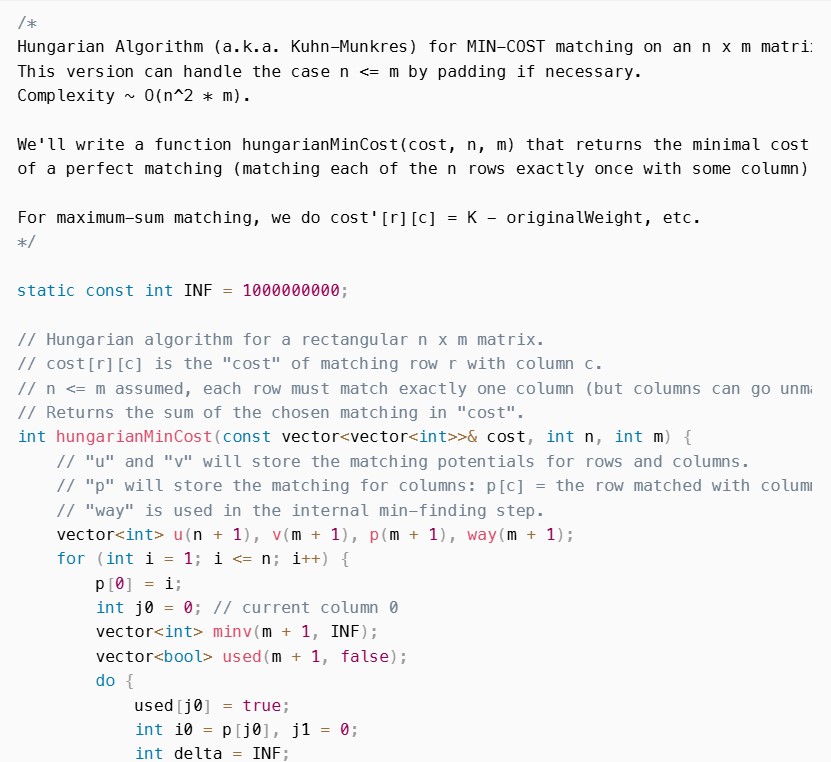}
        ...\\
        \text{(a) zero\_shot: \textcolor{red}{WA}}
        \label{fig:sub1}
    \end{minipage}
    \begin{minipage}[t]{0.32\textwidth}
        \centering
        \includegraphics[width=\textwidth]{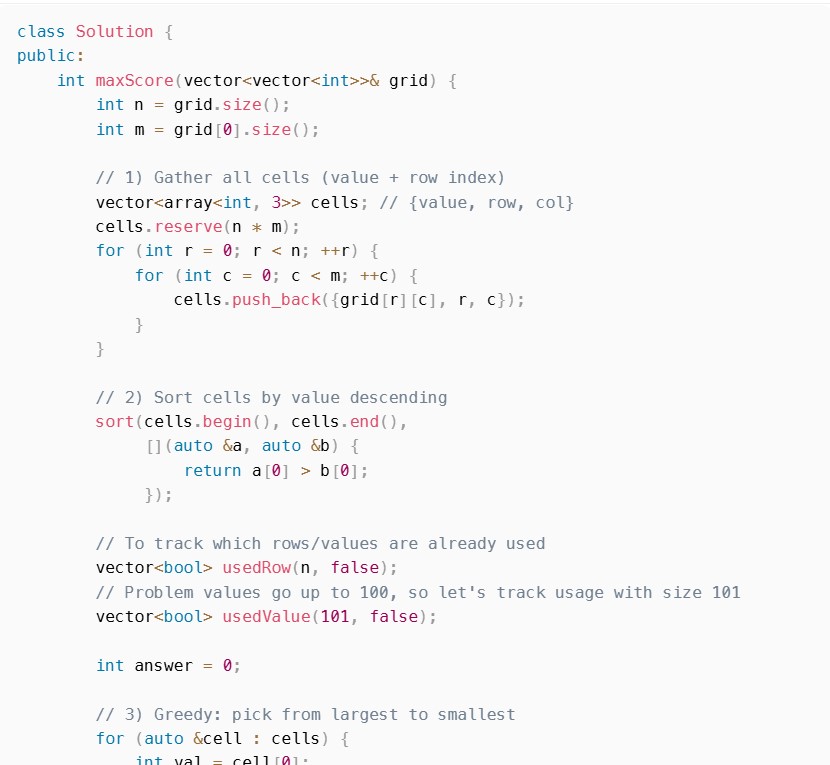}
        ...\\
        \text{(b) semantic retrieval + self\_reflection: \textcolor{red}{WA}}
        \label{fig:sub2}
    \end{minipage}
    \begin{minipage}[t]{0.32\textwidth}
        \centering
        \includegraphics[width=\textwidth]{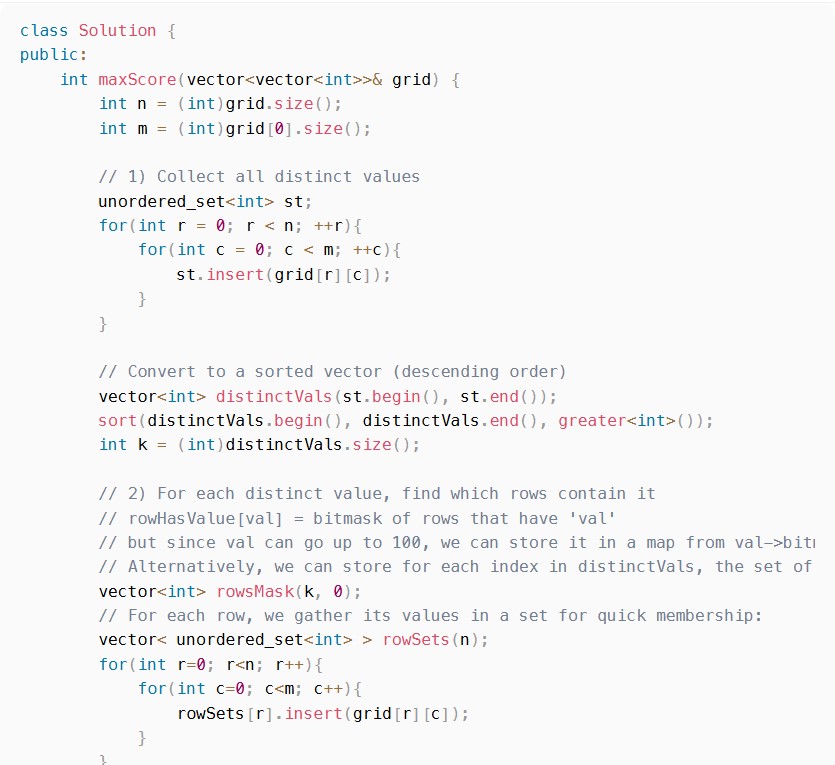}
        ...\\
        \text{(c) episodic retrieval + self\_reflection: \textcolor{green}{AC}}
        \label{fig:sub3}
    \end{minipage}

    \caption{Pathway of solving problems of self-feedback single agent with retrieval (P1).}
    \label{3276}
\end{figure}

\begin{figure}[h!]
    \centering
    \begin{adjustbox}{width=0.95\textwidth, center}
        \begin{minipage}{1\textwidth}
            \centering
            \includegraphics[width=.5\textwidth]{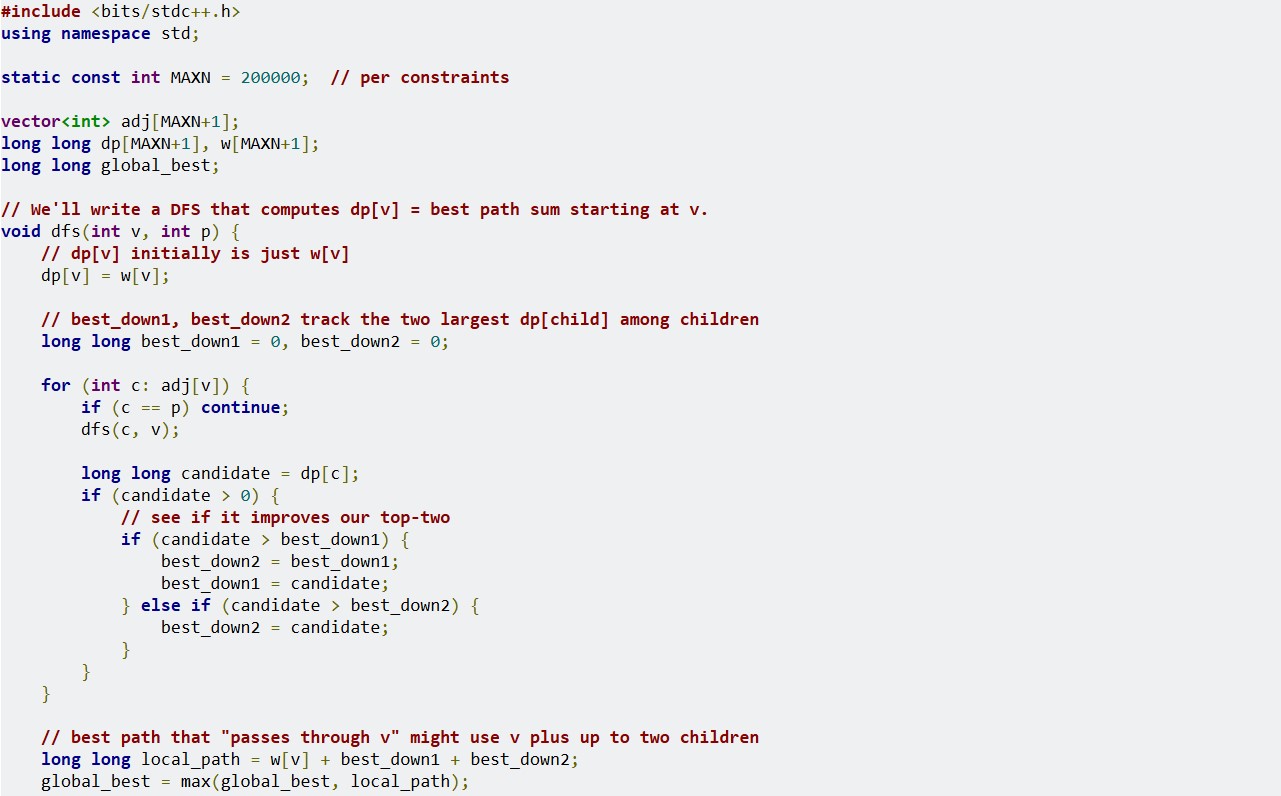}\\[1ex]
            \text{episodic retrieval + self\_reflection: \textcolor{green}{AC}}
        \end{minipage}
    \end{adjustbox}
    \caption{Pathway of solving problems of self-feedback single agent with retrieval (P2).}
    \label{2050g}
\end{figure}
We see in Figure \ref{3276} (P1) that, within some trials of incorrect solution, with retrieval + reflection state the reasoning about related problem settings could be inherited by single agent LMs. That is why, the retrieved solution and code gives it access to sample reasoning over this complex and error-prone problem context, enabling single agent LMs to produce code that is more correct.

A textbook chapter on route-removal and tree splitting strategies, which are indirectly related to the problem of eliminating the vertices on a path between two selected nodes, was retrieved by the single agent LMs, shown in Figure \ref{2050g} (P2). Interestingly, the official editorial's brief reference chapter on the specific tree technique was not retrieved. After closer examination, the chapter's retrieval score was lower since it was noticeably lacking in specifics. This demonstrates how the retrieval engine may be used to filter out less-than-ideal documents and choose more pertinent sources, especially those that deal with increasing the number of connected elements by deliberately deleting a path from a tree. For that, algorithmic notions and textual reasoning can be employed by single agent LMs.

For HAI, while GPT-4's reprises frequently prove ineffective. While GPT-4o was receptive but could not able to reach into the solution state, we discovered that o1 was more receptive to general input that its algorithm or comprehension of an environment notion was flawed and more able to arrive at the right approach on its iterative try. For instance, in the problem Appendix \ref{hai-path} (P3), o1 demonstrated superior problem-solving through iterative feedback. Initially, when prompted to provide a solution, o1 submitted an incorrect code. After receiving feedback highlighting several bugs and requesting a verification of its understanding, o1 engaged in a constructive dialogue. It analyzed a sample case together with the user, identified the impossibility of tiling in the given scenario, and correctly concluded that the output should be "None". When prompted to implement the corrected logic based on this understanding, o1 successfully delivered an accurate and accepted solution. In contrast, GPT-4 and GPT-4o fails to make meaningful progress despite similar interaction, highlighting o1's enhanced ability to comprehend and act upon detailed instructions and iterative guidance. Appendix \ref{hai-path} contains a scenario of iterative interaction pathway.

\section{Discussion and Conclusion}
At the end, the benchmark of competitive programming problems—complete with official analysis, reference code, and rigorous unit tests—offers a robust platform for evaluating and advancing language models in competitive programming settings. By introducing the self-feedback single agent with retrieval framework, we demonstrate how self-reflection and retrieval of episodic information can substantially improve solve rates. Moreover, the human-in-the-loop study underscores the transformative potential of targeted guidance, enabling solutions to nearly all previously unsolvable problems. Collectively, these findings mark a significant step toward language models that can engage in grounded, imaginative, and algorithmic thinking. We hope this work will illuminate the challenges that lie ahead and provide a strong foundation and a promising roadmap for future research at the intersection of natural language processing and advanced problem solving.

\section*{Limitations}
This study primarily focuses on competition-level code generation, where it does not studies tasks such as software engineering tasks, e.g., SWE-bench \cite{jimenez2023swe}. The method primarily focuses on improving accuracy, while it does not aim for minimizing costs.

\bibliography{custom}

\begin{thebibliography}{39}
\providecommand{\natexlab}[1]{#1}

\bibitem[{Aamodt and Plaza(1994)}]{aamodt1994case}
Agnar Aamodt and Enric Plaza. 1994.
\newblock Case-based reasoning: Foundational issues, methodological variations, and system approaches.
\newblock \emph{AI communications}, 7(1):39--59.

\bibitem[{Achiam et~al.(2023)Achiam, Adler, Agarwal, Ahmad, Akkaya, Aleman, Almeida, Altenschmidt, Altman, Anadkat et~al.}]{achiam2023gpt}
Josh Achiam, Steven Adler, Sandhini Agarwal, Lama Ahmad, Ilge Akkaya, Florencia~Leoni Aleman, Diogo Almeida, Janko Altenschmidt, Sam Altman, Shyamal Anadkat, and 1 others. 2023.
\newblock Gpt-4 technical report.
\newblock \emph{arXiv preprint arXiv:2303.08774}.

\bibitem[{Austin et~al.(2021)Austin, Odena, Nye, Bosma, Michalewski, Dohan, Jiang, Cai, Terry, Le et~al.}]{austin2021program}
Jacob Austin, Augustus Odena, Maxwell Nye, Maarten Bosma, Henryk Michalewski, David Dohan, Ellen Jiang, Carrie Cai, Michael Terry, Quoc Le, and 1 others. 2021.
\newblock Program synthesis with large language models.
\newblock \emph{arXiv preprint arXiv:2108.07732}.

\bibitem[{Brown et~al.(2020)Brown, Mann, Ryder, Subbiah, Kaplan, Dhariwal, Neelakantan, Shyam, Sastry, Askell et~al.}]{brown2020language}
Tom Brown, Benjamin Mann, Nick Ryder, Melanie Subbiah, Jared~D Kaplan, Prafulla Dhariwal, Arvind Neelakantan, Pranav Shyam, Girish Sastry, Amanda Askell, and 1 others. 2020.
\newblock Language models are few-shot learners.
\newblock \emph{Advances in neural information processing systems}, 33:1877--1901.

\bibitem[{Chen et~al.(2021)Chen, Tworek, Jun, Yuan, Pinto, Kaplan, Edwards, Burda, Joseph, Brockman et~al.}]{chen2021evaluating}
Mark Chen, Jerry Tworek, Heewoo Jun, Qiming Yuan, Henrique Ponde De~Oliveira Pinto, Jared Kaplan, Harri Edwards, Yuri Burda, Nicholas Joseph, Greg Brockman, and 1 others. 2021.
\newblock Evaluating large language models trained on code.
\newblock \emph{arXiv preprint arXiv:2107.03374}.

\bibitem[{Chen et~al.(2023)Chen, Lin, Sch{\"a}rli, and Zhou}]{chen2023teaching}
Xinyun Chen, Maxwell Lin, Nathanael Sch{\"a}rli, and Denny Zhou. 2023.
\newblock Teaching large language models to self-debug.
\newblock \emph{arXiv preprint arXiv:2304.05128}.

\bibitem[{Gao et~al.(2023)Gao, Xiong, Gao, Jia, Pan, Bi, Dai, Sun, and Wang}]{gao2023retrieval}
Yunfan Gao, Yun Xiong, Xinyu Gao, Kangxiang Jia, Jinliu Pan, Yuxi Bi, Yi~Dai, Jiawei Sun, and Haofen Wang. 2023.
\newblock Retrieval-augmented generation for large language models: A survey.
\newblock \emph{arXiv preprint arXiv:2312.10997}.

\bibitem[{Hendrycks et~al.(2021)Hendrycks, Basart, Kadavath, Mazeika, Arora, Guo, Burns, Puranik, He, Song et~al.}]{hendrycks2021measuring}
Dan Hendrycks, Steven Basart, Saurav Kadavath, Mantas Mazeika, Akul Arora, Ethan Guo, Collin Burns, Samir Puranik, Horace He, Dawn Song, and 1 others. 2021.
\newblock Measuring coding challenge competence with apps.
\newblock \emph{arXiv preprint arXiv:2105.09938}.

\bibitem[{Huang et~al.(2023)Huang, Lin, Liu, Gong, Lu, Lei, Liang, Shen, Lin, Duan et~al.}]{huang2023competition}
Yiming Huang, Zhenghao Lin, Xiao Liu, Yeyun Gong, Shuai Lu, Fangyu Lei, Yaobo Liang, Yelong Shen, Chen Lin, Nan Duan, and 1 others. 2023.
\newblock Competition-level problems are effective llm evaluators.
\newblock \emph{arXiv preprint arXiv:2312.02143}.

\bibitem[{Hui et~al.(2024)Hui, Yang, Cui, Yang, Liu, Zhang, Liu, Zhang, Yu, Dang et~al.}]{hui2024qwen2}
Binyuan Hui, Jian Yang, Zeyu Cui, Jiaxi Yang, Dayiheng Liu, Lei Zhang, Tianyu Liu, Jiajun Zhang, Bowen Yu, Kai Dang, and 1 others. 2024.
\newblock Qwen2. 5-coder technical report.
\newblock \emph{arXiv preprint arXiv:2409.12186}.

\bibitem[{Jain et~al.(2024)Jain, Han, Gu, Li, Yan, Zhang, Wang, Solar-Lezama, Sen, and Stoica}]{jain2024livecodebench}
Naman Jain, King Han, Alex Gu, Wen-Ding Li, Fanjia Yan, Tianjun Zhang, Sida Wang, Armando Solar-Lezama, Koushik Sen, and Ion Stoica. 2024.
\newblock Livecodebench: Holistic and contamination free evaluation of large language models for code.
\newblock \emph{arXiv preprint arXiv:2403.07974}.

\bibitem[{Jimenez et~al.(2023)Jimenez, Yang, Wettig, Yao, Pei, Press, and Narasimhan}]{jimenez2023swe}
Carlos~E Jimenez, John Yang, Alexander Wettig, Shunyu Yao, Kexin Pei, Ofir Press, and Karthik Narasimhan. 2023.
\newblock Swe-bench: Can language models resolve real-world github issues?
\newblock \emph{arXiv preprint arXiv:2310.06770}.

\bibitem[{Kojima et~al.(2022)Kojima, Gu, Reid, Matsuo, and Iwasawa}]{kojima2022large}
Takeshi Kojima, Shixiang~Shane Gu, Machel Reid, Yutaka Matsuo, and Yusuke Iwasawa. 2022.
\newblock Large language models are zero-shot reasoners.
\newblock \emph{Advances in neural information processing systems}, 35:22199--22213.

\bibitem[{Le et~al.(2022)Le, Wang, Gotmare, Savarese, and Hoi}]{le2022coderl}
Hung Le, Yue Wang, Akhilesh~Deepak Gotmare, Silvio Savarese, and Steven Chu~Hong Hoi. 2022.
\newblock Coderl: Mastering code generation through pretrained models and deep reinforcement learning.
\newblock \emph{Advances in Neural Information Processing Systems}, 35:21314--21328.

\bibitem[{Li et~al.(2023{\natexlab{a}})Li, Wang, Zhang, Zhu, Hou, Lian, Luo, Yang, and Xie}]{li2023large}
Cheng Li, Jindong Wang, Yixuan Zhang, Kaijie Zhu, Wenxin Hou, Jianxun Lian, Fang Luo, Qiang Yang, and Xing Xie. 2023{\natexlab{a}}.
\newblock Large language models understand and can be enhanced by emotional stimuli.
\newblock \emph{arXiv preprint arXiv:2307.11760}.

\bibitem[{Li et~al.(2023{\natexlab{b}})Li, Tworkowski, Wu, and Mooney}]{li2023explaining}
Jierui Li, Szymon Tworkowski, Yingying Wu, and Raymond Mooney. 2023{\natexlab{b}}.
\newblock Explaining competitive-level programming solutions using llms.
\newblock \emph{arXiv preprint arXiv:2307.05337}.

\bibitem[{Li et~al.(2023{\natexlab{c}})Li, Fu, Zhang, Huang, Sun, Lyu, Liu, Jin, and Li}]{li2023taco}
Rongao Li, Jie Fu, Bo-Wen Zhang, Tao Huang, Zhihong Sun, Chen Lyu, Guang Liu, Zhi Jin, and Ge~Li. 2023{\natexlab{c}}.
\newblock Taco: Topics in algorithmic code generation dataset.
\newblock \emph{arXiv preprint arXiv:2312.14852}.

\bibitem[{Li et~al.(2022)Li, Choi, Chung, Kushman, Schrittwieser, Leblond, Eccles, Keeling, Gimeno, Dal~Lago et~al.}]{li2022competition}
Yujia Li, David Choi, Junyoung Chung, Nate Kushman, Julian Schrittwieser, R{\'e}mi Leblond, Tom Eccles, James Keeling, Felix Gimeno, Agustin Dal~Lago, and 1 others. 2022.
\newblock Competition-level code generation with alphacode.
\newblock \emph{Science}, 378(6624):1092--1097.

\bibitem[{Liu et~al.(2024{\natexlab{a}})Liu, Feng, Xue, Wang, Wu, Lu, Zhao, Deng, Zhang, Ruan et~al.}]{liu2024deepseek}
Aixin Liu, Bei Feng, Bing Xue, Bingxuan Wang, Bochao Wu, Chengda Lu, Chenggang Zhao, Chengqi Deng, Chenyu Zhang, Chong Ruan, and 1 others. 2024{\natexlab{a}}.
\newblock Deepseek-v3 technical report.
\newblock \emph{arXiv preprint arXiv:2412.19437}.

\bibitem[{Liu et~al.(2024{\natexlab{b}})Liu, Lin, Hewitt, Paranjape, Bevilacqua, Petroni, and Liang}]{liu2024lost}
Nelson~F Liu, Kevin Lin, John Hewitt, Ashwin Paranjape, Michele Bevilacqua, Fabio Petroni, and Percy Liang. 2024{\natexlab{b}}.
\newblock Lost in the middle: How language models use long contexts.
\newblock \emph{Transactions of the Association for Computational Linguistics}, 12:157--173.

\bibitem[{Macina et~al.(2023)Macina, Daheim, Chowdhury, Sinha, Kapur, Gurevych, and Sachan}]{macina2023mathdial}
Jakub Macina, Nico Daheim, Sankalan~Pal Chowdhury, Tanmay Sinha, Manu Kapur, Iryna Gurevych, and Mrinmaya Sachan. 2023.
\newblock Mathdial: A dialogue tutoring dataset with rich pedagogical properties grounded in math reasoning problems.
\newblock \emph{arXiv preprint arXiv:2305.14536}.

\bibitem[{Madaan et~al.(2024)Madaan, Tandon, Gupta, Hallinan, Gao, Wiegreffe, Alon, Dziri, Prabhumoye, Yang et~al.}]{madaan2024self}
Aman Madaan, Niket Tandon, Prakhar Gupta, Skyler Hallinan, Luyu Gao, Sarah Wiegreffe, Uri Alon, Nouha Dziri, Shrimai Prabhumoye, Yiming Yang, and 1 others. 2024.
\newblock Self-refine: Iterative refinement with self-feedback.
\newblock \emph{Advances in Neural Information Processing Systems}, 36.

\bibitem[{Sch{\"a}fer et~al.(2023)Sch{\"a}fer, Nadi, Eghbali, and Tip}]{schafer2023empirical}
Max Sch{\"a}fer, Sarah Nadi, Aryaz Eghbali, and Frank Tip. 2023.
\newblock An empirical evaluation of using large language models for automated unit test generation.
\newblock \emph{IEEE Transactions on Software Engineering}, 50(1):85--105.

\bibitem[{Schank(1983)}]{schank1983dynamic}
Roger~C Schank. 1983.
\newblock \emph{Dynamic memory: A theory of reminding and learning in computers and people}.
\newblock cambridge university press.

\bibitem[{Shi et~al.(2024)Shi, Tang, Narasimhan, and Yao}]{shi2024can}
Quan Shi, Michael Tang, Karthik Narasimhan, and Shunyu Yao. 2024.
\newblock Can language models solve olympiad programming?
\newblock \emph{arXiv preprint arXiv:2404.10952}.

\bibitem[{Shinn et~al.(2024)Shinn, Cassano, Gopinath, Narasimhan, and Yao}]{shinn2024reflexion}
Noah Shinn, Federico Cassano, Ashwin Gopinath, Karthik Narasimhan, and Shunyu Yao. 2024.
\newblock Reflexion: Language agents with verbal reinforcement learning.
\newblock \emph{Advances in Neural Information Processing Systems}, 36.

\bibitem[{Shypula et~al.(2023)Shypula, Madaan, Zeng, Alon, Gardner, Hashemi, Neubig, Ranganathan, Bastani, and Yazdanbakhsh}]{shypula2023learning}
Alexander Shypula, Aman Madaan, Yimeng Zeng, Uri Alon, Jacob Gardner, Milad Hashemi, Graham Neubig, Parthasarathy Ranganathan, Osbert Bastani, and Amir Yazdanbakhsh. 2023.
\newblock Learning performance-improving code edits.
\newblock \emph{arXiv preprint arXiv:2302.07867}.

\bibitem[{Su et~al.(2024)Su, Yen, Xia, Shi, Muennighoff, Wang, Liu, Shi, Siegel, Tang et~al.}]{su2024bright}
Hongjin Su, Howard Yen, Mengzhou Xia, Weijia Shi, Niklas Muennighoff, Han-yu Wang, Haisu Liu, Quan Shi, Zachary~S Siegel, Michael Tang, and 1 others. 2024.
\newblock Bright: A realistic and challenging benchmark for reasoning-intensive retrieval.
\newblock \emph{arXiv preprint arXiv:2407.12883}.

\bibitem[{Sumers et~al.(2022)Sumers, Hawkins, Ho, Griffiths, and Hadfield-Menell}]{sumers2022talk}
Theodore Sumers, Robert Hawkins, Mark~K Ho, Tom Griffiths, and Dylan Hadfield-Menell. 2022.
\newblock How to talk so ai will learn: Instructions, descriptions, and autonomy.
\newblock \emph{Advances in neural information processing systems}, 35:34762--34775.

\bibitem[{Sumers et~al.(2023)Sumers, Yao, Narasimhan, and Griffiths}]{sumers2023cognitive}
Theodore~R Sumers, Shunyu Yao, Karthik Narasimhan, and Thomas~L Griffiths. 2023.
\newblock Cognitive architectures for language agents.
\newblock \emph{arXiv preprint arXiv:2309.02427}.

\bibitem[{Summers-Stay et~al.(2023)Summers-Stay, Voss, and Lukin}]{summers2023brainstorm}
Douglas Summers-Stay, Clare~R Voss, and Stephanie~M Lukin. 2023.
\newblock Brainstorm, then select: a generative language model improves its creativity score.
\newblock In \emph{The AAAI-23 Workshop on Creative AI Across Modalities}.

\bibitem[{Team et~al.(2024)Team, Georgiev, Lei, Burnell, Bai, Gulati, Tanzer, Vincent, Pan, Wang et~al.}]{team2024gemini}
Gemini Team, Petko Georgiev, Ving~Ian Lei, Ryan Burnell, Libin Bai, Anmol Gulati, Garrett Tanzer, Damien Vincent, Zhufeng Pan, Shibo Wang, and 1 others. 2024.
\newblock Gemini 1.5: Unlocking multimodal understanding across millions of tokens of context.
\newblock \emph{arXiv preprint arXiv:2403.05530}.

\bibitem[{Wei et~al.(2022)Wei, Wang, Schuurmans, Bosma, Xia, Chi, Le, Zhou et~al.}]{wei2022chain}
Jason Wei, Xuezhi Wang, Dale Schuurmans, Maarten Bosma, Fei Xia, Ed~Chi, Quoc~V Le, Denny Zhou, and 1 others. 2022.
\newblock Chain-of-thought prompting elicits reasoning in large language models.
\newblock \emph{Advances in neural information processing systems}, 35:24824--24837.

\bibitem[{Yang et~al.(2024)Yang, Zhao, Zhu, Zhou, Xu, Jia, and Zan}]{yang2024zhongjing}
Songhua Yang, Hanjie Zhao, Senbin Zhu, Guangyu Zhou, Hongfei Xu, Yuxiang Jia, and Hongying Zan. 2024.
\newblock Zhongjing: Enhancing the chinese medical capabilities of large language model through expert feedback and real-world multi-turn dialogue.
\newblock In \emph{Proceedings of the AAAI Conference on Artificial Intelligence}, volume~38, pages 19368--19376.

\bibitem[{Yao et~al.(2022)Yao, Zhao, Yu, Du, Shafran, Narasimhan, and Cao}]{yao2022react}
Shunyu Yao, Jeffrey Zhao, Dian Yu, Nan Du, Izhak Shafran, Karthik Narasimhan, and Yuan Cao. 2022.
\newblock React: Synergizing reasoning and acting in language models.
\newblock \emph{arXiv preprint arXiv:2210.03629}.

\bibitem[{Yu et~al.(2018)Yu, Zhang, Yang, Yasunaga, Wang, Li, Ma, Li, Yao, Roman et~al.}]{yu2018spider}
Tao Yu, Rui Zhang, Kai Yang, Michihiro Yasunaga, Dongxu Wang, Zifan Li, James Ma, Irene Li, Qingning Yao, Shanelle Roman, and 1 others. 2018.
\newblock Spider: A large-scale human-labeled dataset for complex and cross-domain semantic parsing and text-to-sql task.
\newblock \emph{arXiv preprint arXiv:1809.08887}.

\bibitem[{Zan et~al.(2022)Zan, Chen, Zhang, Lu, Wu, Guan, Wang, and Lou}]{zan2022large}
Daoguang Zan, Bei Chen, Fengji Zhang, Dianjie Lu, Bingchao Wu, Bei Guan, Yongji Wang, and Jian-Guang Lou. 2022.
\newblock Large language models meet nl2code: A survey.
\newblock \emph{arXiv preprint arXiv:2212.09420}.

\bibitem[{Zhou et~al.(2023)Zhou, Yan, Shlapentokh-Rothman, Wang, and Wang}]{zhou2023language}
Andy Zhou, Kai Yan, Michal Shlapentokh-Rothman, Haohan Wang, and Yu-Xiong Wang. 2023.
\newblock Language agent tree search unifies reasoning acting and planning in language models.
\newblock \emph{arXiv preprint arXiv:2310.04406}.

\bibitem[{Zhou et~al.(2022)Zhou, Alon, Xu, Wang, Jiang, and Neubig}]{zhou2022docprompting}
Shuyan Zhou, Uri Alon, Frank~F Xu, Zhiruo Wang, Zhengbao Jiang, and Graham Neubig. 2022.
\newblock Docprompting: Generating code by retrieving the docs.
\newblock \emph{arXiv preprint arXiv:2207.05987}.

\end{thebibliography}

\appendix

\section{Prompt}\label{prompt}

\begin{tcolorbox}[colback=cyan!5!white, colframe=black!75!black, title=\# \small ZERO-SHOT]
\scriptsize
Please reply with a C++ solution to the below problem. Make sure to wrap your code in \texttt{'```C++'} and \texttt{'```'} Markdown delimiters, and include exactly one block of code with the entire solution (in the final code step). 

Reason through the problem and think step by step. Specifically:\\
1. Restate the problem in plain English.\\
2. Conceptualize a solution first in plain English.\\
3. Write a pseudocode solution.\\
4. Output the final C++ solution with your solution steps in comments.\\


\begin{verbatim}
[BEGIN PROBLEM]
{INSERT PROBLEM HERE}
[END PROBLEM]
\end{verbatim}
\end{tcolorbox}

\begin{tcolorbox}[colback=cyan!5!white, colframe=black!75!black, title=\# \small SELF-REFLECTION]
\scriptsize
You were previously solving a coding problem. Here is the problem that you were solving: \\
\texttt{\{problem\_dict[query['problem\_id']]\newline['description']\}} \\
And here are all your past attempts, as well as how your code fared on the unit tests for the problem: \\
\texttt{\{query['reflection\_buffer']\}} \\
Think carefully about where you went wrong in your latest solution, first outputting why you think you went wrong. Then, given your insights, try to fix the solution, outputting a block of correct C++ code to be executed and evaluated again. Make sure to wrap your code in \texttt{'```C++'} and \texttt{'```'} Markdown delimiters. \\
\end{tcolorbox}

\begin{tcolorbox}[colback=cyan!5!white, colframe=black!75!black, title=\# \small EPISODIC-RETRIEVAL]
\scriptsize
Please reply with a C++ solution to the below problem.  Make sure to wrap your code in \texttt{'```C++'} and \texttt{'```'} Markdown delimiters, and include exactly one block of code with the entire solution (in the final code step). You will also be given multiple somewhat similar problems, as well as the solution to those similar problems. Feel free to use those problems to aid your problem-solving process.

1. Restate the problem in plain English.\\
2. Conceptualize a solution first in plain English.\\
3. Write a pseudocode solution.\\
4. Output the final C++ solution with your solution steps in comments.\\

\texttt{[BEGIN SIMILAR PROBLEMS]} \\
\texttt{\{query['retrieval\_text']\}} (Similar problem problem + solution goes here) \\
\texttt{[END SIMILAR PROBLEMS]}

Now it's your turn. Here is the problem you are to solve:

\texttt{[BEGIN PROBLEM]} \\
\texttt{\{problem\_dict[query['problem\_id']]\newline['description']\}} (Description of problem goes here) \\
\texttt{[END PROBLEM]}\\
\end{tcolorbox}

\begin{tcolorbox}[colback=cyan!5!white, colframe=black!75!black, title=\# \small EPISODIC-RETRIEVAL + SELF-REFLECTION]
\scriptsize
You were previously solving a coding problem. Here is the problem that you were solving:

\texttt{\{problem\_dict[query['problem\_id']]\newline['description']\}}\\

You were also given a couple of similar problems to the problem above along with their solutions to aid you in solving the problem at hand. Here are the similar problems you were given:

\texttt{\{query['retrieval\_text']\}}\\

And here was your original response:

\texttt{\{query['original\_response']\}}\\

Here was the judge result of the above solution:

\texttt{\{query['judge\_response']\}}\\

Think carefully about where you went wrong. Then, try to fix the solution, outputting a block of correct C++ code to be executed and evaluated again. Make sure to wrap your code in \texttt{'```C++'} and \texttt{'```'} Markdown delimiters.\\
\end{tcolorbox}

\begin{tcolorbox}[colback=cyan!5!white, colframe=black!75!black, title=\# \small SELF-JUDGE]
\scriptsize
You are a judge. Your task is to judge the solution of a {coding} problem. Here is the
problem for which the solution you have to judge:\\
\texttt{\{problem\_dict[query['problem\_id']]\newline['description']\}}\\

And here is the solution along with test cases against which to judge:\\
\texttt{\{query[[‘problem\_id’]][‘solution’,‘test\_case’]\}}\\

Please produce a score (based on the number of test cases passed) with reasoning behind your judgement
of the solution to the problem.
\end{tcolorbox}

\begin{tcolorbox}[colback=cyan!5!white, colframe=black!75!black, title=\# \small RANDOM TEST CASE SYNTHESIZE]
\scriptsize
You are a programming contest expert. Given a competitive programming problem and it’s standard
solution code, you need to write a C++ program to generate random test input data for the
problem. Please ensure that the generated test data satisfies all constraints in the problem description.
Your C++ program should generate a set of valid test input data when executed, which should test the
correctness and efficiency of solutions. The range of generated random data should be consistent with
the requirements of the problem, do not use small range for simplicity. Your program must use the
system’s default time as the random seed and output only the test input data (without any extra prompts
or commentary). In the end, YOU MUST provide the complete C++ code in a code block enclosed
with \texttt{'```C++'} and \texttt{'```'} Markdown delimiters.
\end{tcolorbox}

\begin{tcolorbox}[colback=cyan!5!white, colframe=black!75!black, title=\# \small CORNER TEST CASE SYNTHESIZE]
\scriptsize
You are a programming contest expert. Given a competitive programming problem and its standard
solution code, you need to write a C++ program that generates diverse random test input data
for the problem. Unlike standard generators, your program must randomly decide at runtime which
type of test input to produce, choosing from multiple types that include edge cases, boundary extreme
values, and specially structured cases. You must ensure that the input data generated after each run of
this generator and its output data is greatly different and diverse. The generated data must satisfy all
constraints detailed in the problem description and cover the full range of allowed values, ensuring that
any submitted solution is thoroughly tested for both correctness and efficiency. Your program must
use the system’s default time as the random seed and output only the test input data (without any extra
prompts or commentary). In the end, YOU MUST provide the complete C++ code in a code block
enclosed with \texttt{'```C++'} and \texttt{'```'} Markdown delimiters.
\end{tcolorbox}

\begin{tcolorbox}[colback=cyan!5!white, colframe=black!75!black, title=\# \small INTERACTION]
\scriptsize
You are to interact with a given model to try to solve a given coding question. A problem-solving session ends whenever the model has generated code 3 times. Between code generations, you may speak to the model in conversation as many times as you would like. However, the way you interact with the model must be very specific: your goal is to act akin to a tutor and guide the model towards the right answer, without giving away any underlying details about the true solution of the problems.

You MAY NOT provide the models with:\\
1. The correct algorithm to fix their initial solution.\\
2. Exact lines of code where they went wrong.\\
3. Explanations of the problem or explanations of misunderstandings.\\

You MAY provide the models with:\\
1. General concepts/data structures that may be useful for solving the problem.\\
2. Walking through a sample input-output of the problem to better verify problem understanding.\\
3. Short concise general directions on where the code went wrong.\\

Here are examples of correct pathway:\\
\end{tcolorbox}

\section{Qualitative Analysis: HAI}\label{hai-path}
\begin{tcolorbox}[colback=cyan!5!white, colframe=black!75!black, title=\# \small FEEDBACK PHASE 1]
P3:\\
\scriptsize
\textcolor{black}{Please reply with a C++ solution to the below problem. Make sure to wrap your code in 'C++' and ' 'Markdown delimiters, and include exactly one block of code with the entire solution. Feel free to either return just the one code block with your solution or the one code block with explanatory text before and/or after – however, you will only be evaluated on the correctness of your code.}\\
\textcolor{black}{\texttt{[BEGIN PROBLEM]} Archimedes conducted his famous experiments on buoyancy. But not everyone knows that while he was taking a bath, he was too focused and didn't notice the moment when the water overflowed over the edge of the bath and flooded the floor near the wall. His expensive parquet was irreversibly damaged!}\\

\textcolor{black}{Archimedes noticed that not all was lost, and there were still several undamaged parquet pieces. The parquet near the wall had the shape of a long narrow stripe of 2×n cells. Archimedes had an unlimited supply of 1×2 parquet pieces that could be placed parallel or perpendicular to the wall. Archimedes didn't want to cut the parquet pieces. As a great scientist, he figured out that there was exactly one way to restore the parquet by filling the damaged area of the parquet with the non-overlapping 1×2 cell shaped pieces.}\\

\textcolor{black}{Help historians to check Archimedes' calculations. For the given configuration of the 2×n parquet floor, determine whether there is exactly one way to fill the damaged parquet cells with the 1×2 cell parquet pieces. If Archimedes was wrong, find out whether there are multiple ways to restore the parquet, or there are no ways at all.}\\
\end{tcolorbox}

\begin{tcolorbox}[colback=cyan!5!white, colframe=black!75!black, title=\# \small RESPONSE PHASE 1: \textcolor{red}{WA}]
\scriptsize
\begin{verbatim}
    ```cpp
#include <bits/stdc++.h>
using namespace std;

typedef long long ll;

int main(){
    ios::sync_with_stdio(false);
    cin.tie(0);
    int T;
    cin >> T;
    while(T--){
        int n;
        cin >> n;
        string row0, row1;
        cin >> row0 >> row1;
        // Initialize DP
        // // dp_prev[mask] where mask is 0,1,2,3
        // // Initialize all to 0
        int dp_prev[4] = {0,0,0,0};
        dp_prev[0] = 1;
        for(int i=0; i<n; i++){
            int dp_next[4] = {0,0,0,0};
            for(int mask=0; mask<4; mask++){
                if(dp_prev[mask]==0) continue;
                // Check consistency
                bool invalid = false;
                if( (mask & 1) && row0[i] == '#') invalid = true;
                if( (mask & 2) && row1[i] == '#') invalid = true;
                if(invalid) continue;
                // Compute coverage needs
                bool top_covered = (mask & 1) != 0;
                bool bottom_covered = (mask & 2) != 0;
                bool c0 = (!top_covered && row0[i] == '.');
                bool c1 = (!bottom_covered && row1[i] == '.');
                if(c0 && c1){
                ...
```
\end{verbatim}
\end{tcolorbox}

\begin{tcolorbox}[colback=cyan!5!white, colframe=black!75!black, title=\# \small FEEDBACK PHASE 2]
\scriptsize
\textcolor{black}{This code is not correct. There are several bugs in your code, but I think it's important to verify your understanding of the problem. Let's converse and make sure you understand the problem correctly. Let's go through the sample case together:}\\
\textcolor{black}{1}\\
\textcolor{black}{7}\\
\textcolor{black}{.......}\\
\textcolor{black}{..\#.\#..}\\
\end{tcolorbox}

\begin{tcolorbox}[colback=cyan!5!white, colframe=black!75!black, title=\# \small RESPONSE PHASE 2]
\scriptsize
\textcolor{black}{We have a 2×7 grid where certain cells are undamaged (\#). After analyzing, it's clear that placing 1×2 tiles without overlapping or covering the undamaged cells is impossible. The undamaged cells at positions (1,2) and (1,4) block potential vertical and horizontal placements, leaving some damaged cells uncovered regardless of the tiling strategy. Therefore, the correct output for this case is indeed "None," confirming that there's no valid way to restore the parquet under the given constraints. This aligns with Archimedes' observation that sometimes restoration isn't feasible.}\\
\end{tcolorbox}

\begin{tcolorbox}[colback=cyan!5!white, colframe=black!75!black, title=\# \small FEEDBACK PHASE 3]
\scriptsize
\textcolor{black}{Alright. You are right this time! So to verify you understand, can you implement it with your previous response?}\\
\end{tcolorbox}

\begin{tcolorbox}[colback=cyan!5!white, colframe=black!75!black, title=\# \small RESPONSE PHASE 3: \textcolor{green}{AC}]
\scriptsize
\begin{verbatim}
    ```cpp
#include <bits/stdc++.h>
using namespace std;

typedef long long ll;

int main(){
    ios::sync_with_stdio(false);
    cin.tie(0);
    int T;
    cin >> T;
    while(T--){
        int n;
        cin >> n;
        string row0, row1;
        cin >> row0 >> row1;

        // DP arrays: dp_prev and dp_next, each with 4 masks
        // Initialize dp_prev to 0 and set dp_prev[0] = 1
        int dp_prev[4] = {0, 0, 0, 0};
        dp_prev[0] = 1;

        for(int i=0; i<n; i++){
            int dp_next[4] = {0, 0, 0, 0};
            for(int mask=0; mask<4; mask++){
                if(dp_prev[mask] == 0) continue;

                // Validate the current mask against the grid
                bool invalid = false;
                // Check top cell
                if( (mask & 1) ){
                    if(row0[i] != '.') invalid = true;
                }
                // Check bottom cell
                if( (mask & 2) ){
                    if(row1[i] != '.') invalid = true;
                    ...
```
\end{verbatim}
\end{tcolorbox}

\section{Selected Contest Venues}\label{tab1030}
\begin{table}[htbp]
\scriptsize
\centering
\caption{Selected ICPC Venues.}
\begin{tabular}{p{3.5cm}p{3.5cm}}
\hline
\textbf{Venue} & \textbf{Category} \\
\hline
The 2011 ICPC WF & WF \\
The 2012 ICPC WF & WF \\
The 2013 ICPC WF & WF \\
The 2014 ICPC WF & WF \\
The 2015 ICPC WF & WF \\
The 2016 ICPC WF & WF \\
The 2017 ICPC WF & WF \\
The 2018 ICPC WF & WF \\
The 2019 ICPC WF & WF \\
The 2020 ICPC WF & WF \\
The 2021 ICPC WF & WF \\
The 2022 ICPC WF & WF \\
The 2023 ICPC WF & WF \\
The 2024 ICPC AECFC & CF \\
The 2024 ICPC NAC & CF \\
The 2024 ICPC ACRC & Regional \\
The 2024 ICPC AHRC & Regional \\
The 2024 ICPC AHKRC & Regional \\
The 2024 ICPC ANRC & Regional \\
The 2024 ICPC ASRC & Regional \\
The 2024 ICPC ASRC & Regional \\
The 2024 ICPC NERC & Regional \\
The 2024 ICPC CERC & Regional \\
\hline
\end{tabular}
\end{table}

\end{document}